\documentclass{ecai} 

\usepackage{latexsym}
\usepackage{amssymb}
\usepackage{amsmath}
\usepackage{amsthm}
\usepackage{booktabs}
\usepackage{enumitem}
\usepackage{graphicx}
\usepackage{color}
\usepackage{algorithm} 
\usepackage{algpseudocode} 

\usepackage{multirow}
\usepackage{cleveref}

\newcommand{\argmin}{\mathop{\mathrm{argmin}}}

\newcommand{\BibTeX}{B\kern-.05em{\sc i\kern-.025em b}\kern-.08em\TeX}

\begin{document}


\begin{frontmatter}


\paperid{123} 


\title{How to Protect Models against Adversarial Unlearning?}


\author[A]{\fnms{Patryk}~\snm{Jasiorski}\orcid{0009-0005-0994-8708}\thanks{Corresponding author. Email: patryk.jasiorski@identt.pl}}
\author[B]{\fnms{Marek}~\snm{Klonowski}\orcid{0000-0002-3141-8712}\footnotemark}
\author[B]{\fnms{Michał}~\snm{Woźniak}\orcid{0000-0003-0146-4205}} 

\address[A]{IDENTT Corporation}
\address[B]{Wrocław University of Science and Technology}


\begin{abstract}
AI models need to be unlearned to fulfill the requirements of legal acts such as the AI Act or GDPR, and also because of the need to remove toxic content, debiasing, the impact of malicious instances, or changes in the data distribution structure in which a model works. 
Unfortunately, removing knowledge 
may cause undesirable side effects, such as a deterioration in model performance. In this paper, we investigate the problem of adversarial unlearning, where a malicious party intentionally sends unlearn requests to deteriorate the model's performance maximally. We show that this phenomenon and the adversary's capabilities depend on many factors, primarily on the backbone model itself and strategy/limitations in 
selecting data to be unlearned. The main result of this work is a new method of protecting model performance from these side effects, both in the case of unlearned behavior resulting from spontaneous processes and 
adversary actions.

\end{abstract}

\end{frontmatter}


\section{Introduction}

Unlearning involves removing unwanted knowledge from a trained model or the impact of selected training examples on its behavior. Let us
consider a model $M$ trained on a certain set $D$. We want to remove the influence of a particular subset of the training set $D_{\text{u}}\subset D$ on the model, leaving the rest of the knowledge unaltered.  
That is, the goal of unlearning is to produce a model $M'$ that would be "as if it had been trained" on the set $D_{\text{r}}$.  The most natural way to achieve this goal is to train the model from scratch in $D_{\text{r}}$. This resulting model is denoted as $M_{\text{Gold}}$ and is considered a perfect ("gold") one. However, in practice, generating $M_{\text{Gold}}$ is inacceptably resource intensive.

Recently, many works have tried to perform unlearning much more efficiently. That is, to obtain a model $M'$ that is "the same as"  $M_{\text{Gold}}$ (exact unlearning) or is "close to" $M_{\text{Gold}}$ (approximate unlearning). Note that this "model closeness" can be defined differently (see \cite{nguyen2022surveymachineunlearning,jeong2024sokchallengesopportunitiesfederated}). However, the fundamental problem is that both models $M'$ and $M_{\text{Gold}}$ are randomized structures.  That is, the learning process is usually randomized and, for the same training set, we can obtain different models according to some distribution (usually unknown and impossible to describe in practice). This leads to fundamental problems in assessing model similarity and determining whether specific data have not been learned correctly from a given model instance~\cite{DBLP:conf/uss/ThudiJSP22}.

The problem we focus on in this paper is that unlearning, even the perfect one, causes changes in the model's behavior that may not be desired by the model's owner. 
The unlearning request itself may 
result from one of the many attacks \cite{hu2024dutyforgetrightassured,DBLP:journals/aim/ChenLZH25}. The second observation is that unlearning a model may be the consequence of fulfilling a duty to the owner of the data used in training. In such a case, the model owner must implement 
an unlearned request. On the other hand, it may not want such an unlearned request at all and try to mitigate its effects.

The main contribution of this work is the proposition of a method that can help correct the effects (both malicious and spontaneous) of unlearning. The main idea is to somehow replace one element indicated for removal with another real element "similar" to the one being removed. The intention of this action, which we will call \textit{healing}, is different from performance recovery based on different techniques that range from fine tuning to knowledge destillation methods (see, e.g., \cite{DBLP:conf/sp/CaoJZG23, DBLP:conf/middleware/DhasadeD0KVW24, DBLP:journals/corr/abs-2307-02106}).


In a nutshell, 
the main idea is to create a  
reserve of "spare elements" that will replace the elements to be unlearned. Effective operation requires a careful construction of this set and a policy of replacing elements.

The latter aims to correct the system's operation towards the ideal model without the removed element. Healing is 
to mitigate the effects of undesirable (but required) unlearning. Additionally, we experimentally show that 
using unlearning can lead to significant model degradation even in the simplest scenarios (without elaborated attacks). We also show that in practice models differ significantly in their susceptibility to aberrations resulting from unlearned. We also show that healing has varying effectiveness, which depends on many factors.


\section{Related Works}


Recently, there has been a steady increase in interest in unlearning, which is manifested, among other things, in several review articles on the topic~\cite{nguyen2022surveymachineunlearning,jeong2024sokchallengesopportunitiesfederated,Survey,awesome}. 
The most common \textbf{motivation} for unlearning is to comply with legal regulations (AI ACT, GDPR, CCPA \cite{eu-269-2014, CCPA}) that give users the Right to Be Forgotten (RFTB). Other reasons concern various types of model correction, such as removing toxic/destructive content or elements used to install backdoor attacks~\cite{SecSurv}. Unlearning can also be considered as a method of debiasing~\cite{Chen:2023} or correction of malfunctioning systems (e.g., removing the Clever Hans effect) \cite{CHmedicine, Lapuschkin2019}. Some real-life examples of unlearning cases can be found in~\cite{shaik2024exploringlandscapemachineunlearning} and references therein.

The main requirement that an unlearning algorithm must meet is efficiency in terms of computational complexity, i.e., it consumes only a small amount of computational power needed to train a model from scratch. 
The other requirement is to remove the influence of the indicated elements on the model. The \textbf{evaluation} of this is seemingly simple and compares the updated model with the model learned from scratch. It is assumed that a good unlearning algorithm will make them "similar". However, a similarity can be understood differently, as closeness of parameters or similarity of responses (\cite{DBLP:conf/cvpr/GolatkarARPS21, mercuri2022introductionmachineunlearning}). We may find many approaches to measure the correctness of unlearning, such as the membership inference test (\cite{DBLP:journals/corr/abs-2010-10981}), the information-theoretic approach \cite{DBLP:conf/cvpr/GolatkarAS20}, the approach based on differential privacy (\cite{guo2023certifieddataremovalmachine}), or metrics based on the so-called relearning (e.g. \textit{ anamnesis index} ~\cite{DBLP:journals/tifs/ChundawatTMK23}) to enumerate only a few. 
However, there is no consensus on the use of specific metrics or approaches in this regard. Thudi et al.~\cite{DBLP:conf/uss/ThudiJSP22} showed that most of these definitions are inadequate to some extent and lead to significant aberrations (the same model can be unlearned and unlearned at the same time according to popular definitions). These problems result from the probabilistic nature of the dataset itself and the methods of learning and unlearning. It seems that defining the forgetting measure is less important from the point of view of attacks (including those using unlearning). However, it is crucial from the point of view of evaluating the healing model method (see the discussion in Section~\ref{healing}).



This work introduces \textit{healing}, i.e., using substitute examples to \textbf{protect against adversarial unlearning}. To our knowledge, this approach has not been used before.
However, it is worth noting that many articles consider unlearning as a way of hostile actions. In~\cite{huang2024unlearnburnadversarialmachine}, authors present how the accuracy of the classifier can be substantially worsened by unlearning. However, the presented attack assumes "unlearning" elements that were actually \textbf{not} used in the training set. Note that such an attack can still be practical unless safeguards are used to prevent launching unlearning for elements outside the training set (such as hash-checking). A similar approach is also discussed in \cite{DBLP:conf/kdd/QianZLMH23}, where attacks on models are based on maliciously crafted examples inserted into the training set. The goal is to force an incorrect behavior on the target sample to classify it incorrectly. Another malicious use of unlearning can be found in \cite{DBLP:conf/nips/DiDA0S23}, where the authors show that a carefully crafted element added to the training set can lead to a poisoning attack, which will only work after applying unlearning. Yet another malicious activity related to unlearning is described in the paper \cite{DBLP:journals/corr/abs-2109-08266}, showing how unlearning can increase computations.


Interesting results on other attacks, especially in the context of real-life implementations in the MLaaS model, can be found in \cite{hu2024dutyforgetrightassured} or  \cite{ma2024releasingmalevolencebenevolencemenace}, where the adversary can worsen the models' performance. However, the described strategy requires the adversary to contribute data to the training set.


It is also worth mentioning \textit{Adversarial Machine Unlearning} from \cite{di2025adversarial}, whose main contribution is to integrate membership inference attacks into the design of machine unlearning via the Stackelberg Game.
In this context, "adversarial" refers to the game-theoretic relationship between the unlearner and the auditor that improves the unlearning algorithm. However, this work is quite different from our contribution.

\section{Threats of Adversarial Unlearning}
\label{attack}


This section will introduce the \textit{healing method}, but first, let us make some remarks about the risks associated with unlearning. Even in the case of perfect unlearning, i.e., retraining from scratch, the unlearned model may have worse performance just because it was trained on a smaller data set. In particular, we can observe this phenomenon when the model is under-fitted. Additionally, even occasional approximated unlearning in complex models can destroy the model's accuracy. This problem was immediately noticed, and many unlearning procedures implement various techniques of model performance recovery (\cite{nguyen2022surveymachineunlearning,Survey}) as a kind of post-processing built-in the unlearning algorithm. We also need to consider the \textit{adversarial unlearning} scenario, where a malicious party may intentionally generate an unlearning request to degrade the model's performance maximally. In experimental studies, we show that it is possible to significantly degrade the performance of classifiers even without advanced attacks (e.g.,\cite{huang2024unlearnburnadversarialmachine, DBLP:conf/kdd/QianZLMH23, DBLP:conf/nips/DiDA0S23}).

Threats (both from spontaneous and malicious actions) to a 
machine learning model can be described as a certain algorithm $\mathcal{A}$ (called adversary). The adversary is limited by certain assumptions that reflect access to the attacked system. The following constraints should be taken into account when defining $\mathcal{A}$:


1. \textbf{Number of elements to be unlearned}.

2. \textbf{Fraction of elements the adversary can indicate for removal}. Of course, removing two different elements can have different effects on the performance (e.g., accuracy, but also inference time).

3. \textbf{Knowledge of the system}. We can consider \textit{system-agnostic} (no access to the system), \textit{output-aware} (the adversary has access to results of classification of chosen elements), and \textit{parameter-aware} adversary (the adversary knows the parameters, for example, weights of DNN before and after unlearning).

4. \textbf{Knowledge of the training set.} The adversary may know the training set\textit{known training set} or even have the ability to add some elements to the training set \textit{ selected training set}. if 
the adversary does not know the training set, we will refer to it as \textit{blind} adversary.

The goal is therefore to find an algorithm $\mathcal{A}$ that minimizes Accuracy($\mathcal{A}(M)$) (or its expected value, if $\mathcal{A}$ is randomized), 
considering the adversary's limitations.  Clearly, finding the optimal $\mathcal{A}$ for some scenarios may be impossible in practice.

Note that different models are adequate for various real-life scenarios. For example, the output-aware adversary corresponds to the situation when the attacker can access the black-box model and classify some elements. The parameter-aware adversary corresponds to a glass-box attack when the adversary can access the model as an administrator. A fundamentally different, although also realistic, scenario is with the adversary being able to influence part of the training set (chosen trading set) and from the model, when the adversary only knows part of the training set (known training set). However, 
certain combinations of settings make no practical sense.

Note that the above taxonomy could be richer, for example, when considering the situation where unlearning from the adversary is mixed with spontaneous unlearning from non-malicious users of the system. 
More advanced attacks presented so far assume that precisely crafted elements are presented to be unlearned (a known training set model) or even a selected training set model. 
It seems that despite the impressive technical results, such solutions may not be easy to implement in practice. It is because we assumed that the attacked system is isolated from other (non-adversarial users). In practice, however, if we consider large-scale systems (those that cannot be easily learned from scratch), we must take into account that the adversarial unlearn requests will be blended with spontaneous, non-malicious unlearn requests from regular users. This can strengthen the attack (further reducing the model's knowledge), but can also lead to model stabilization and thwarting the attack. It can be hypothesized that begnin unlearning may \textbf{in some cases} paradoxically protect the system from adversarial unlearning. This requires extensive research and is beyond the scope of our work.

\section{Model Healing}\label{healing}



As we mentioned, unlearning could be a tool for malicious actions that can effectively destroy the system. Thus, this chapter presents \textit{healing}, a method that can mitigate the effects of malicious unlearning. In a nutshell, \textit{healing} retrains the model using a set $z^{*}$ that consists of examples of instances "similar" to the unlearned set $z$ chosen from a "spare" instance set.
Let us emphasize that 

\begin{itemize}[topsep=0pt]
\item \textit{Healing} is \textbf{not} unlearning or even a type of performance recovery after unlearning. Intuitively, unlearning tries to remove the influence of a given element on the model, while healing is a step toward reversing these changes. 
\item \textit{Healing} leads to a certain paradox. On the one hand, we remove some $z$ from the model, and on the other hand, we replace it with another (real, not synthetic) $z^*$. Due to the similarity between $z$ and $z^*$, some properties of $z$ are transferred back to the model. 
\end{itemize}

\paragraph{Healing Algorithms Descriptions}

The proposed solution's main characteristic is how to generate a “spare” instance set. The proposed solution's main characteristic is how to generate a “spare” instance set. Thus, let us present two natural approaches that are appropriate for different settings depending on the nature of the model and data availability. 

\paragraph{General Spare Set}

A set $D_S$ of $k$ elements \textit{ spare elements} is randomly chosen from the original training set. 
$D$, and a model is trained in $D \setminus D_S$. 
$k$ is a parameter that depends on the size of the training set and the expected number of unlearning requests.


When one requests to unlearn an element $z$, we randomly select an instance $z^*$ from $D_s$ that is the \textbf{most similar} to $z$ according to a chosen metric $d$. The chosen $z^*$ will be used for the \textit{healing} procedure and removed from $D_S$.

\begin{algorithm}
	\caption{General Spare Set Healing}  
	\begin{algorithmic}[1]
		\If {$D_s \neq \emptyset $}
		 \State $ z^* \in_{R} \argmin\limits_{z'  \in  D_s } d(z', z^*)$   
		   \If $\mbox{  }d(z, z^*) < \delta $
              \State Use $z^*$  for healing  
               \State  $D_S :=  D_S \setminus \{ z^*\}$   
           \EndIf
		\EndIf 
	\end{algorithmic} 
\end{algorithm}

\paragraph{Twins Strategy}



The Twins strategy involves searching for each element of a training set $z$ its "twin" $z^*$, i.e., $d(z,z^*)\leq \delta$. A spare element (twin) $z^*$  will be used for the healing if $z$ is requested to be unlearned. Such a strategy requires a large amount of additional data. However, it can only be used for a part of the training set that is particularly important or exposed to unlearning requests (e.g., limited rights to $z$). One may also use a strategy in which some data is included in the training set only when we have a surrogate of that data to replace the original data in advance.
We may naturally extend this strategy to triplets, quadruplets, etc. More generally, each element used for training is associated with a subset of $k$ similar elements - surrogates, used as needed.

\paragraph{Appropriate  similarity measure}

The basic issue is properly selecting the similarity assessment method for the elements we use as surrogates. This comes down to the proper selection of the metric function $d$. This function must be adapted to the data type and capture the natural similarity properties of the elements. In the experimental part, we used Euclidean distance directly for the compared elements, i.e., $ d_{\ell_2}(z,z') = |z - z' |_2$  (e.g., raw pixel-wise). 
We also conducted experiments in feature space using cosine similarity:

\begin{equation*}\label{eq:cosine}
        \cos\!\bigl(f(z),\,f(z')\bigr) =\frac{f(z)\cdot f(z')}{\|f(z)\|_2~\|f(z')\|_2},
    \end{equation*}
\noindent
and Mahalanobis distance:
    \begin{equation*}\label{eq:mahalanobis}
        d_M\bigl(f(z), f(z')\bigr) = \sqrt{
          \bigl(f(z) - f(z')\bigr)^\top
          \Sigma^{-1}
          \bigl(f(z) - f(z')\bigr)},
    \end{equation*}
    \noindent
    where $\Sigma$ is the covariance of the features embedded in $f(z')$ over $D_{\text{r}}$.

A natural extension of this approach is appropriate to craft a function $d$ that could accurately represent similarity and define the substitutability of elements $z$ and $z’$  after unlearning.
The appropriate approach depends on the details of the model, but such a solution may be based on Siamese neural networks.

\section{Experimental Results}~\label{experimental}  

This section presents experimental results for different configurations of backbone models, data sets and unlearning strategies. We also consider different adversarial settings and various healing strategies.   
 
\subsection{Experimental Setup}~\label{D&B}
   We conducted experiments on three datasets: MNIST~\cite{Lecun1998}, CIFAR‑10~\cite{krizhevsky2009learning}, and AFHQ~\cite{choi2020starganv2}. MNIST comprises $70 000$ grayscale images of handwritten digits (28×28 px), CIFAR‑10 contains $60 000$ color images (32×32 px) across $10$ classes, and AFHQ offers $16 130$ high‑quality animal‑face images (512×512 px) in three domains (cats, dogs, wildlife). Each dataset is paired with an appropriate backbone: a small custom CNN for MNIST, a pre-trained ResNet‑50 for CIFAR‑10, and a pretrained EfficientNet‑B0 for AFHQ.

     We perform unlearning using four methods. 
            \paragraph{\mbox{     }Naive}
            Model retraining from scratch.
            \paragraph{\mbox{   }SISA}
                (Sharded, Isolated, Sliced, and Aggregated) proposed by Bourtoule ~\cite{Bourtoule:2021} as a more efficient exact unlearning method than naive. It follows a simple idea: the training set is partitioned into $k$ subsets of similar sizes.  A model is trained independently on each subset. Then, the $k$ models are used independently. The response of the entire system is the aggregated response of $k$ models (e.g., the response in classification can be the most common response over $k$ responses). The gain in unlearning is obvious: updating a single element does not require training on the entire dataset because removing a single element changes only one model. We therefore only need to perform $1/k$ computations for what we have in naive unlearning.
            \paragraph{\mbox{   }Fisher Unlearning}
                Fisher Unlearning, proposed in~\cite{Golatkar:2020} is an approximate unlearning strategy that takes advantage of the Fisher Information Matrix (FIM) to estimate how parameters should be updated to forget specific training samples. The Fisher information matrix $F(\theta)$ measures how sensitive the model predictions are to small parameter changes $\theta$. It is formally defined as~\cite{martens2020newinsightsperspectivesnatural}:
                        
                        \begin{equation*}\label{eq:fim}
                             F(\theta) = \mathbb{E}_{(x,y) \sim D}\left[\nabla_\theta \log p(y|x,\theta)\,\nabla_\theta \log p(y|x,\theta)^\top\right].
                        \end{equation*}

                Fisher Unlearning updates the weights in a single Newton‑style step, followed by a calibrated Fisher‑noise injection:
                        
                        \begin{equation*}\label{eq:fisher}
                             {
                             \theta' = \theta - F(\theta)^{-1} \nabla_\theta L(\theta; D_{\text{remaining}}) + \sigma F(\theta)^{-1/4} b,
                             }
                        \end{equation*}
                         with $b \sim \mathcal{N}(0, I)$ and $\sigma > 0$ as a noise parameter.
                         
                         Regarding the original paper, for deep learning models, the update simplifies to only noise injection. Practically, calculating the full inverse of the Fisher matrix is computationally expensive. Thus, efficient approximations such as diagonal or Kronecker-factored approximations (KFAC) \cite{martens2020optimizingneuralnetworkskroneckerfactored} are employed. 
         \paragraph{\mbox{   }Influence Unlearning}
         Influence Unlearning~\cite{koh2020understandingblackboxpredictionsinfluence}  attempts to approximate unlearning employing the concept of influence function, which quantifies the effect of individual training points on the model parameters.  Considering a model with parameters $\theta$ trained on a dataset $D$ 
         by minimizing a twice‑differentiable loss $L(x,y;\theta)$, the local curvature is captured by the empirical Hessian  
        \begin{equation*}\label{eq:hessian}
             H(\theta)=\frac{1}{N}\sum_{(x,y)\in D}\nabla_{\theta}^{2}L(\theta;x,y).
        \end{equation*}
        
         If a single point $z^{\ast}=(x^{\ast},y^{\ast})$ was to be unlearned, the parameters would shift by  
        \begin{equation*}\label{eq:single-if}
             \Delta\theta(z^{\ast})=-\,H(\theta)^{-1}\,\nabla_{\theta}L(\theta; x^{\ast},y^{\ast}),
        \end{equation*}
        
         When an entire batch $D_{\text{u}}\subset D$ must be forgotten, influences add linearly, giving the one‑shot update  
        \begin{equation*}\label{eq:many-if}
             \theta'=\theta- H(\theta)^{-1}\!\sum_{(x,y)\in D_{\text{remove}}}\nabla_{\theta}L(\theta; x,y).
        \end{equation*}
        
         Because storing or inverting the full Hessian is impractical, especially in deep learning scenarios, a common approach is to approximate inverse‑Hessian products with LiSSA iterations~\cite{Agarwal2016SecondOrderSO} or a short conjugate‑gradient solve~\cite{10.5555/3104322.3104416}.

 \paragraph{Unleraning procedures}
  We begin by training a baseline model $M$ and its SISA counterpart $M_{SISA}$. Then, for each set $D_{\text{u}}\subset D$ to be unlearned we set
$D_{\text{r}} = D \setminus D_{\text{u}}$  and perform following procedures to obtain four unlearened models: 
        \begin{description}
        \item[Naive:]  gold model $M'_{Gold}$ is trained  from scratch on $D_{\text{remaining}}$.
          \item[SISA] $M'_{SISA}$ is computed from $M_{SISA}$ using the standard SISA procedure.    
          \item[Fisher:] $M'_{Fisher}$ is computed from $M$ using an approximation of the Fisher Information Matrix.
          \item[Influence:] $M'_{Influnece}$ is calculated from the inverse Hessian gradient product $M$ using the LiSSA recursion.
         \end{description}

The experimental results and conclusions are highly dependent on the specific implementations and selection of methods and parameters. It should be emphasized that the procedures used are standard.

In our initial experiments, we remove, at random, 5\%, 10\%, 20\% and 30\% of the training samples, dividing each deletion set into minibatches and applying sequential unlearning updates as in~\cite{mahadevan2021certifiablemachineunlearninglinear}. Base model $M$ was trained for N epochs – 10 for MNIST, 5 for CIFAR-10, and 3 for AFHQ. $M_{Gold}$ was also trained  for N epochs. $M_{SISA}$ had a structure of 3 shards, further sliced into 5 slices. Adam optimizer was used with a learning rate of $10^{-3}$.

\subsection{Basic  Results}
Below we present selected results from experiments on unlearning for different models of adversary. In the next part, we will relate them to the healing procedure.

        \begin{table}[!ht]
        \centering
        \caption{Accuracy and training time for each method on the full datasets.}
        \label{tab:baseline_sisa_compact}
        \begin{tabular}{llcc}
        \toprule
        Dataset   & Method              & Accuracy\footnotemark[1] & Time (s) \\
        \midrule
        \multirow{2}{*}{MNIST}    & Baseline (CNN)      & 0.9898                   & 167       \\
                                   & SISA                & 0.9891                   & 147       \\
        \midrule
        \multirow{2}{*}{CIFAR‑10} & Baseline (ResNet‑50) & 0.9527                   & 1782      \\
                                   & SISA                & 0.9463                   & 1745      \\
        \midrule
        \multirow{2}{*}{AFHQ}     & Baseline (EffNet‑B0) & 0.9980                   & 510       \\
                                   & SISA                & 0.9973                   & 445       \\
        \bottomrule
        \end{tabular}
        \end{table}
        
        \footnotetext[1]{Since all three benchmarks are approximately class‑balanced and our models’ precision, recall, and $F_{1}$‑scores track overall accuracy, we omit those redundant metrics here for conciseness.}

        ~\Cref{tab:baseline_sisa_compact} reports the predictive metrics and training time for each model before any unlearning.  SISA achieves a 12–15\% reduction in runtime compared to standard training, with only a marginal drop (under 0.7pp) in accuracy across all three datasets.

        \begin{figure}[t]
          \centering
          \includegraphics[width=\linewidth]{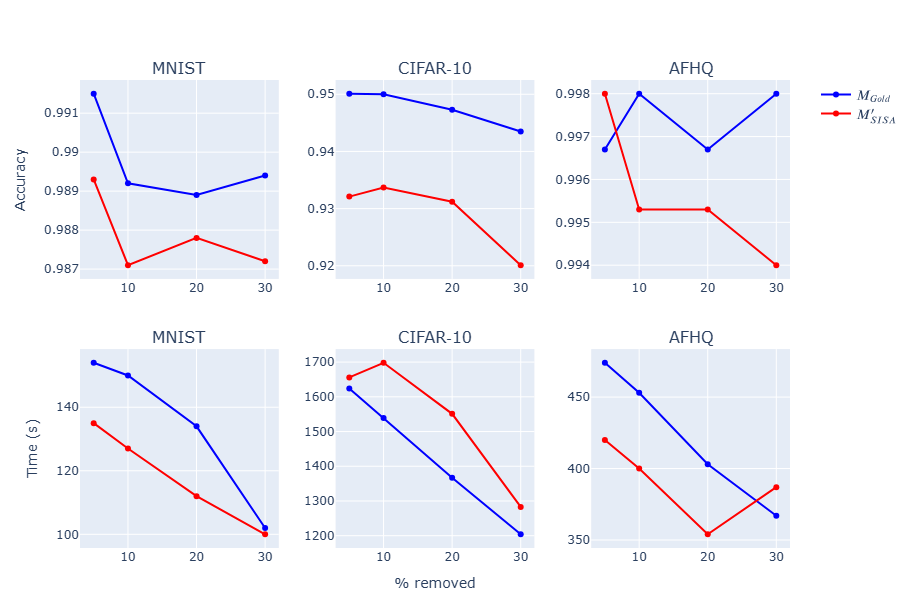}
          \caption{Test accuracy (top) and unlearning latency (bottom) of the full‑retrain model $M_{Gold}$ (blue) versus the shard‑wise update $M'_{SISA}$ (red), as the fraction of deleted samples increases from 5\% to 30\%.}
          \label{fig:unlearning_plots}
        \end{figure}

        \begin{figure}[t]
            \centering
            \includegraphics[width=\linewidth]{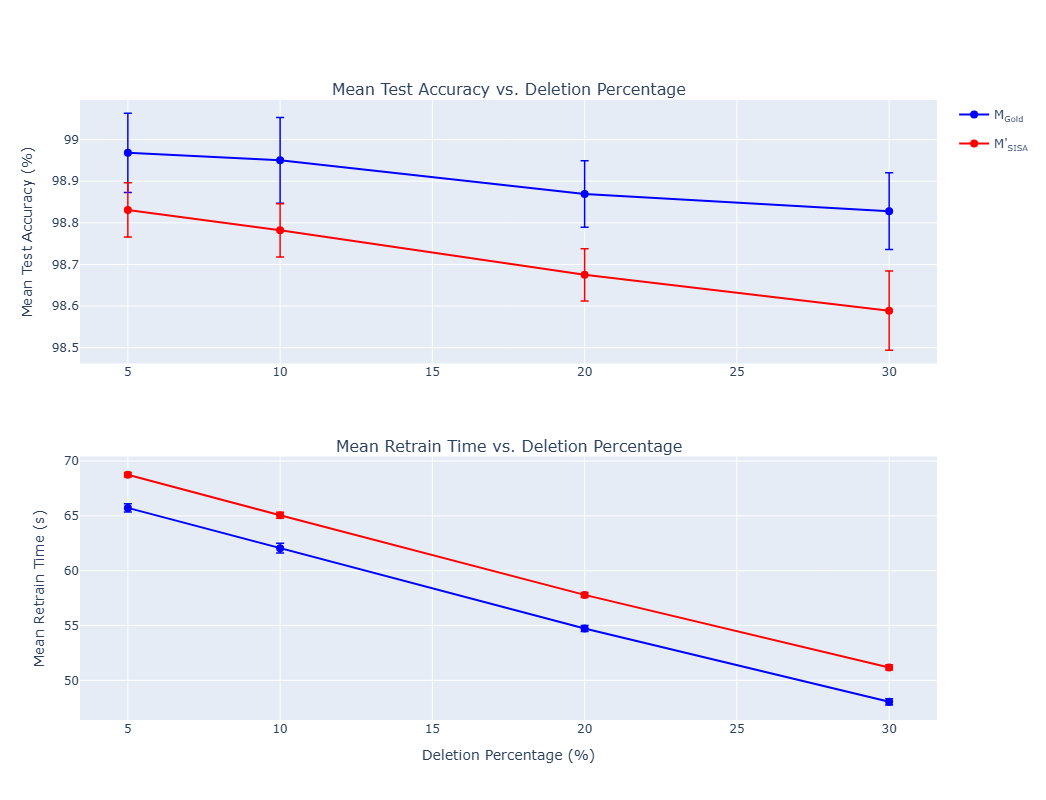}
            \caption{Test accuracy (top) and unlearning latency (bottom) of the full‑retrain model $M_{Gold}$ (blue) versus the shard‑wise update $M'_{SISA}$ (red), as the fraction of deleted samples increases from 5\% to 30\%, evaluated on the MNIST for 10 independent runs.}
            \label{fig:acc_10runs}
        \end{figure}
        
        ~\Cref{fig:unlearning_plots}, relative to the baselines in ~\Cref{tab:baseline_sisa_compact}, shows that $M_{Gold}$ experiences negligible accuracy drop even as up to 30\% of the training set is removed, while $M'_{SISA}$ accuracy declines steadily with larger deletion fractions, reflecting the weakening of its shard‑wise submodels. Focusing on the MNIST dataset over 10 runs, as depicted in ~\Cref{fig:acc_10runs}, we can observe more stable results concerning model accuracy degradation. Unlearning time for $M'_{SISA}$ stays within a small margin of full‑retrain time even under the pessimistic assumption that every shard must be updated, so the real runtime cost is generally lower.

        \begin{table}[ht]
        \centering
        \caption{Summary of Fisher unlearning on the baseline models: test accuracy and unlearning time for each fraction removed.}
        \label{tab:fisher_summary}
        \begin{tabular}{lccc}
        \toprule
        Dataset      & \% Removed & Accuracy & Time (s) \\
        \midrule
        \multirow{4}{*}{MNIST}   & 5\%   & 10.1\% &  43 \\
                                 & 10\%  & 14.3\% & 131 \\
                                 & 20\%  & 10.1\% & 495 \\
                                 & 30\%  & 12.9\% &  93 \\
        \addlinespace
        \multirow{4}{*}{CIFAR‑10}& 5\%   & 10.0\% & 548 \\
                                 & 10\%  & 10.0\% &1055 \\
                                 & 20\%  & 10.0\% &1478 \\
                                 & 30\%  & 10.0\% & 877 \\
        \addlinespace
        \multirow{4}{*}{AFHQ}    & 5\%   & 33.5\% & 695 \\
                                 & 10\%  & 33.3\% & 217 \\
                                 & 20\%  & 33.3\% & 394 \\
                                 & 30\%  & 33.3\% & 456 \\
        \bottomrule
        \end{tabular}
        \end{table}

\begin{table*}[ht!]
    \centering
    \caption{Influence unlearning hyperparameters, achieved accuracy and unlearning time for each removal fraction.}
    \label{tab:influence_app}
    \begin{tabular}{lccccccccc}
    \toprule
    \multirow{2}{*}{Dataset} & \multirow{2}{*}{\% Removed} & \multirow{2}{*}{Accuracy} & \multirow{2}{*}{Time [s]} & \multirow{2}{*}{Batch$_r$} & \multirow{2}{*}{Batch$_f$} & \multirow{2}{*}{$\epsilon$}  & \multirow{2}{*}{Max Norm} & \multirow{2}{*}{LiSSA Depth} & \multirow{2}{*}{Scale} \\
                              &                           &                            &                           &                            &                            &                       &                            &                  &       \\
    \midrule
    \multirow{4}{*}{MNIST}   
        & 5\%  & 0.9777 & 15  & 2048 & 1024 & 1e-4 & 1   & 10 & 10   \\
        & 10\% & 0.9553 & 32  & 2048 & 1024 & 1e-4 & 1   & 10 & 10   \\
        & 20\% & 0.9510 & 30  & 2048 & 2048 & 1e-4 & 1   & 10 & 10   \\
        & 30\% & 0.9410 & 45  & 2048 & 2048 & 1e-4 & 1   & 10 & 10   \\

    \addlinespace
    \multirow{17}{*}{CIFAR‑10} 
        & \multirow{6}{*}{5\%}  
            & 0.1648 & 358 & 128 & 256  & 1e-2 & 5   & 7  & 75    \\
        &                       & 0.2742 & 283 & 128 & 256  & 1e-2 & 3   & 5  & 75    \\
        &                       & 0.7646 & 256 & 128 & 256  & 1e-2 & 2   & 5  & 75    \\
        &                       & 0.8829 & 283 & 128 & 256  & 1e-2 & 1   & 5  & 75    \\
        &                       & 0.7660 & 237 & 128 & 256  & 1e-2 & 0.5 & 5  & 75    \\
        &                       & 0.6921 & 275 & 128 & 256  & 1e-2 & 1   & 5  & 500   \\

    \addlinespace
        & \multirow{5}{*}{10\%} 
            & 0.9402 & 111  & 128 & 512  & 1e-3 & 0.5 & 5  & 100   \\
        &                       & 0.5198 & 260  & 128 & 512  & 1e-3 & 5   & 5  & 100   \\
        &                       & 0.8598 & 256  & 128 & 512  & 1e-3 & 1   & 5  & 100   \\
        &                       & 0.8033 & 419  & 128 & 256  & 1e-3 & 1   & 5  & 100   \\
        &                       & 0.8415 & 1055 & 128 & 128  & 1e-3 & 1   & 5  & 100   \\

    \addlinespace
        & \multirow{5}{*}{20\%} 
            & 0.9336 & 82   & 128 & 1024 & 1e-3 & 1   & 1  & 100   \\
        &                       & 0.9353 & 92   & 128 & 1024 & 1e-3 & 1   & 3  & 100   \\
        &                       & 0.8153 & 274  & 128 & 1024 & 1e-3 & 1   & 5  & 100   \\
        &                       & 0.8751 & 368  & 128 & 1024 & 1e-3 & 1   & 10 & 100   \\
        &                       & 0.8745 & 600  & 128 & 1024 & 1e-3 & 1   & 15 & 100   \\

    \addlinespace
        & 30\% & 0.5089 & 143 & 128 & 2048 & 1e+3 & 10  & 10 & 1000  \\

    \bottomrule
    \end{tabular}
\end{table*}

         ~\Cref{tab:fisher_summary} shows the effect of a diagonal‑FIM Fisher update applied to our original baseline model $M$ with up to 30\% random deletion. Even with only 5\% of the data removed the model consistently collapses (test accuracy drops to ~10–33\% - a single‐class output) and stays at that level no matter how we adjust the unlearning batch sizes, noise scale $\sigma$, damping parameter $\epsilon$ or gradient clipping bound. Unlearning times vary with batch$_r$/batch$_f$ with even some exceeding retraining scenario if the batch$_f$ is too small, but none can prevent collapse. This behavior shows that Fisher update without any subsequent fine‑tuning is too harsh under random removal and seems not to be practical in its raw form.
        ~\Cref{tab:influence_app} summarizes Influence unlearning on the original baseline models $M$ across MNIST, CIFAR‑10 (AFHQ results are omitted: on our hardware, any reasonable mini‑batch choice drives unlearning time above the retraining time). On MNIST, Influence degrades from 97.8\% (5\% removal) to 94.1\% (30\% removal) in just 15–45 s, compared to 102–154 s for full retraining ($\approx$3×–10× speedup), but leaves a 2–5 point accuracy gap to $M_{Gold}$. On CIFAR‑10, with carefully chosen hyperparameters, Influence can drop accuracy to 94.0\% in 111s versus 1539s for retraining ($\approx$14× speedup). Other hyperparameter choices yield accuracy as low as 16\%–52\%, and even intermediate settings only reach 76\%–88\%. Such middling results may reflect stronger removal of deleted‐data influence but come at the expense of erasing genuine model information. Overall, Influence unlearning completes in about 1.5× to 6.5× faster than the need for training of $M_{Gold}$.


\subsection{Experiments with Healing Algorithms}
    \label{recover_idea}

    This subsection discusses the experimental results for selected settings of \textit{healing} algorithms.
    Model healing addresses the “knowledge gap” when unlearning procedures remove or alter parameters to forget specific training samples, causing accuracy drops or increased uncertainty on the remaining data. In our experiments, both Influence and Fisher unlearning show such degradation, recovering 94\%–96\% on MNIST and highly variable performance on CIFAR‑10 (16\%–94\%), which motivates the need for a targeted healing step. When the retained dataset $D_{\text{r}}$ is available, the most straightforward remedy is to fine‑tune the unlearned model on it for a few epochs, realigning parameters without reintroducing the specific removed information. This post‑unlearning retraining typically costs only a fraction of full retraining and restores most of the lost accuracy, provided $D_{\text{r}}$ remains large and representative.

    A more advanced healing strategy selects a small set of “twin” replacement samples that semantically cover the regions left behind by the deleted points.

\subsubsection{Experimental Healing Setup}\label{recovery_eksp}
Having observed that unlearning techniques can struggle (\Cref{experimental}), we designed a focused set of experiments to evaluate strategies for healing model performance after unlearning. For this study, we used the same datasets and model pairs (\Cref{D&B}). However, instead of removing large random portions, we simulated a more targeted request: forgetting just $25$ specific images belonging to a single chosen class within the main training data. We first divided the original full training dataset for each benchmark. About $70\%$ was set aside as the primary training set, which we used to build our main models. The remaining $30\%$ formed a separate backup set, kept aside solely to provide potential replacement samples later during the healing phase.

\begin{itemize}[topsep=0pt]
    \item We start by training baseline model $M$ on the primary training for N epochs – specifically, 10 for MNIST, 5 for CIFAR-10, and 3 for AFHQ. Then, we create $M_{Gold}$ by retraining a fresh model from scratch using only the primary training data minus the 25 images, also for N epochs. 
    \item We apply two approximate unlearning methods on $M$ to remove the 25 target samples. Since these methods can have variable outcomes, we ran each one 5 times. We selected the resulting model state closest to the mean accuracy to represent starting points for evaluating healing.
    \item Potential replacement samples for the 25 targeted samples are identified by searching within the replacement set using four different similarity metrics: raw pixel L2 distance, raw pixel Mahalanobis distance (excluded for AFHQ dataset), feature cosine similarity, and feature Mahalanobis distance (based on embeddings extracted from $M$).
    \item We test up to five different data setups for the healing fine-tuning: (i) using the remaining primary training data combined with the method-based twins;
(ii) using only the remaining primary training data with no twins added.
    \end{itemize}

Five healing setups were tested under two conditions: a very brief fine-tuning of just $1$ epoch and a longer fine-tuning duration of $\left\lceil \frac{N}{2} \right\rceil$ epochs - corresponding to half the original baseline training epochs. All healing fine-tuning used the Adam optimizer with a learning rate of $10^{-3}$ and was evaluated against the same validation set derived from $D_{\text{r}}$.

\subsubsection{Experimental Healing Results}\label{recovery_results}

We report the best and worst test accuracies achieved across the five data configurations for each starting model and epoch setting. We compare them against the baseline model $M $ and the target Gold Standard $M_{Gold}$. 
It is important to remember that achieving good healing results relies on having a reasonable starting model produced by the initial approximate unlearning step. This reality means methods' hyperparameters – such as Fisher's noise level ($\sigma$) or the settings for the iterative Hessian approximation algorithm used in Influence - must be chosen carefully through experimentation. Finding the right balance is key: the settings must be effective enough to demonstrate some measurable impact of unlearning, yet not so aggressive that they cause the model to collapse completely, a frequent outcome with less cautious choices. Even with a thoughtful setup, the stochastic nature of these methods means that the results can vary from run to run, which led us to use representative mean-outcome runs as the starting points for our healing evaluations.

\begin{table*}[ht]
\centering
\caption{Model Healing Performance (\%) - Best and Worst Outcomes vs. Benchmarks. BL=Baseline, GS=Gold Standard. FS/IF Init Acc = Accuracy of representative Fisher/Influence model \textit{before} healing. Heal FS/IF columns show results \textit{after} applying healing methods starting from the respective representative model. \textbf{Best Acc}: Highest accuracy across the 5 data configs (4x Remain+Twins, 1x RemainOnly) for the specified epoch count; method achieving best noted in parentheses (R=RemainOnly, T=Best Twin Variant). \textit{Worst Acc}: Lowest accuracy across the 5 data configs for that epoch count.
\textit{$\Delta pp~Best-GS$}: Difference between Best Acc and Gold Standard.}
\label{tab:healing_summary}
\resizebox{\textwidth}{!}{
\begin{tabular}{l l | c c | c c | c c | c c}
\toprule
\textbf{Dataset} & \textbf{Metric} & \textbf{BL} & \textbf{GS} & \textbf{FS Init*} & \textbf{IF Init*} & \textbf{Heal FS (1 Epoch)} & \textbf{Heal FS (N/2 Epochs)} & \textbf{Heal IF (1 Epoch)} & \textbf{Heal IF (N/2 Epochs)} \\
\textbf{(Model)} &                 & \textbf{Acc (\%)} & \textbf{Acc (\%)} & \textbf{Acc (\%)} & \textbf{Acc (\%)} & \textbf{Acc (\%)}       & \textbf{Acc (\%)}   & \textbf{Acc (\%)}       & \textbf{Acc (\%)}   \\
\midrule
MNIST       & \textbf{Best Acc}   & 99.02 & 98.87 & 84.36 & 90.16 & 98.73 (R)            & \textbf{98.98} (T-FeatMahal) & 98.97 (T-RawMahal) & \textbf{99.03} (T-FeatMahal) \\
(CNN)       & \textit{Worst Acc}    & ---   & ---   & ---   & ---   & \textit{98.41}       & \textit{98.50}          & \textit{98.69}       & \textit{98.76}          \\
            & \textit{$\Delta pp$ Best-GS}  & +0.15 & 0.00  & -14.51& -8.71 & -0.14                & +0.11                   & +0.10                & +0.16                   \\
\midrule
CIFAR-10    & \textbf{Best Acc}   & 91.51 & 90.97 & 66.76 & 80.33 & 90.13 (T-RawL2)        & \textbf{91.41} (T-RawL2)      & 90.95 (T-RawMahal)   & \textbf{91.21} (R)            \\
(ResNet50)  & \textit{Worst Acc}    & ---   & ---   & ---   & ---   & \textit{89.55}       & \textit{90.36}          & \textit{89.51}       & \textit{88.13}          \\
            & \textit{$\Delta pp$ Best-GS}  & +0.54 & 0.00  & -29.29& -10.67& -0.84                & +0.44                   & -0.02                & +0.24                   \\
\midrule
AFHQ        & \textbf{Best Acc}   & 99.33 & 99.40 & 96.87 & 91.67 & 99.47 (R)            & \textbf{99.67} (T-FeatCos)  & 99.67 (T-FeatMahal)  & \textbf{99.73} (R)            \\
(EffNetB0)  & \textit{Worst Acc}    & ---   & ---   & ---   & ---   & \textit{99.33}       & \textit{99.27}          & \textit{99.20}       & \textit{98.53}          \\
            & \textit{$\Delta pp$ Best-GS}  & -0.07 & 0.00  & -3.55 & -8.13 & +0.07                & +0.27                   & +0.27                & +0.33                   \\
\bottomrule
\multicolumn{10}{l}{(*) Note: FS/IF Init Acc based on representative run closest to mean; mean/std dev available in \Cref{tab:approx_mean}.} \\
\multicolumn{10}{l}{N/2 Epochs corresponds to 5 (MNIST), 3 (CIFAR-10), and 2 (AFHQ).}
\end{tabular}
}
\end{table*}

\begin{table*}[ht]
\centering
\caption{Model Healing Performance (\%) on MNIST (worst 25 logits) — Remain+Twins vs. Random Replacement.  
BL=Baseline, GS=Gold Standard. 
FS/IF Init Acc = Accuracy of representative Fisher/Influence model \textit{before} healing. Heal FS/IF columns show results \textit{after} applying healing methods starting from respective representative model.  Top block: Remain+Twins variants.  
Bottom block: Random-25 replacement.  
\textbf{Best Acc}: Highest accuracy across methods.  
\textit{Worst Acc}: Lowest accuracy across the Remain+Twins configs.
\textit{$\Delta$ Best–GS}: Difference between Best Acc and Gold Standard.}
\label{tab:mnist_worst25_combined}
\resizebox{\textwidth}{!}{%
\begin{tabular}{l l | c  c  | c   c  | c   c   | c   c}
\toprule
\textbf{Dataset} & \textbf{Metric}
  & \textbf{BL}  & \textbf{GS}
  & \textbf{FS Init*} & \textbf{IF Init*}
  & \textbf{Heal FS (1 Epoch)} & \textbf{Heal FS (5 Epochs)}
  & \textbf{Heal IF (1 Epoch)} & \textbf{Heal IF (5 Epochs)} \\

\textbf{(Model)} &                 & \textbf{Acc (\%)} & \textbf{Acc (\%)} & \textbf{Acc (\%)} & \textbf{Acc (\%)} & \textbf{Acc (\%)}       & \textbf{Acc (\%)}   & \textbf{Acc (\%)}       & \textbf{Acc (\%)}   \\
\midrule
\multirow{6}{*}{MNIST}  
  & \textbf{Best Acc}
    & 99.02 & 98.72
    & 78.19 & 11.71
    & \textbf{98.71} (T-FeatMahal)
    & \textbf{98.90} (T-RawL2)
    & \textbf{74.73} (T-RawMahal)
    & 87.73 (T-RawMahal) \\
  & \textit{Worst Acc}
    & ---     & ---
    & ---     & ---
    & \textit{98.18}
    & \textit{98.67}
    & \textit{66.13}
    & \textit{80.22} \\
  & \textit{$\Delta$ Best–GS}
    & +0.15      & 0.00
    & -20.53     & -87.01
    & -0.01
    & +0.18
    & -23.99
    & -10.99 \\
\cmidrule(lr){2-10}
  & \textbf{Best Acc}
    & 99.02 & 98.72
    & 78.19 & 11.71
    & 98.40 (T-Random)
    & 98.49 (T-Random)
    & 65.70 (T-Random)
    & \textbf{92.65} (T-Random) \\
  & \textit{$\Delta$ Best–GS}
    & +0.15      & 0.00
    & -20.53     & -87.01
    & -0.32
    & -0.23
    & -33.02
    & -5.37 \\
\bottomrule
\multicolumn{10}{l}{(*) FS/IF Init Acc based on representative run closest to the mean of worst-25-logits run (\Cref{tab:approx_mean}).}
\end{tabular}%
}
\end{table*}

\begin{table*}[ht!]
\centering
\caption{Mean performance over 5 runs when forgetting 25 samples from a target class for Fisher and Influence method.}
\label{tab:approx_mean}
\begin{tabular}{@{} c c c c c c c c c @{}}
\toprule
Dataset & Method & Target Class & Mean Acc & StdDev Acc & Mean Time (s) & $\epsilon$ & {$\sigma$ / LiSSA Depth} & Scale \\
\midrule

\multirow{2}{*}{MNIST}    & Fisher & 7      & 0.8428 & 0.0444 & 7  & 1.00e-06 & 3.50e-03 & {-} \\
         & Influence & 7      & 0.9089 & 0.0124 & 30 & 1.00e-06 & 5000     & 1000 \\
\midrule

\multirow{2}{*}{CIFAR-10} & Fisher & deer   & 0.6168 & 0.2792 & 207& 1.00e-06 & 3.50e-05 & {-} \\
         & Influence & deer   & 0.8030 & 0.0012 & 520& 1.00e-06 & 1000     & 10000 \\
\midrule

\multirow{2}{*}{AFHQ}     & Fisher & wildlife& 0.9585 & 0.0330 & 88 & 1.00e-06 & 1.00e-03 & {-} \\
         & Influence & wildlife& 0.9127 & 0.0689 & 93 & 1.00e-06 & 100      & 100 \\

\midrule

\multirow{2}{*}{MNIST (worst 25 logits)}     & Fisher & various & 0.8237 & 0.0794 & 7 & 1.00e-06 & 3.50e-03 & {-} \\
         & Influence & various & 0.1089 & 0.2110 & 30 & 1.00e-06 & 5000      & 1000 \\

\bottomrule
\end{tabular}
\end{table*}

After analyzing the results, several key trends emerged regarding the effectiveness of model healing. Firstly, and most importantly, targeted fine-tuning demonstrably recovers performance significantly. Even when starting from severely degraded states, such as the representative Fisher model on CIFAR-10, healing methods could restore accuracy close to or exceeding the Gold Standard level.
Comparing the fine-tuning durations, extending the healing from $1$ epoch to $N/2$ epochs yields better results. Across all datasets and starting points (FS or IF), the best accuracy achieved after $N/2$ epochs was consistently higher than the best accuracy after just $1$ epoch. This suggests that while a single epoch provides rapid improvement, allowing the model more adaptation time on the healing data leads to a more complete recovery, often closing the final gap to the Gold Standard performance. For instance, on CIFAR-10, starting from the Influence model, $1$-epoch healing achieved the best accuracy just below Gold (-0.02pp), while $3$-epoch healing surpassed it (+0.24pp).
The specific data used for fine-tuning – just the remaining data (R) or the remaining data plus twin samples (T) – also made a difference. Fine-tuning longer, as stated before, generally produced better results overall. Examining these N/2-epoch results that combining the remaining data with carefully chosen twin samples often led to the highest final accuracy (achieving the best outcome in 4 out of 6 start-point/dataset combinations, particularly effective using feature-based twins on MNIST and AFHQ, and raw-L2 twins for Fisher on CIFAR-10). However, simply fine-tuning the remaining data (R) also performed very strongly, yielding the best results in the other two cases (Influence start on CIFAR-10 and AFHQ). For the quicker 1-epoch healing, adding specific twin samples frequently gave the best initial recovery boost, suggesting twins can provide useful corrective information rapidly. The difference between the best and worst accuracy for any given setting was usually not extreme, suggesting the process is reasonably stable. However, the choice of healing data influences the final performance level.

To further investigate healing under more severe degradation, \Cref{tab:mnist_worst25_combined} presents results for an MNIST experiment where initial unlearning specifically targeted the 'worst 25 logits' samples, leading to substantially lower starting model accuracies. This table compares the efficacy of our 'Remain+Twins' healing strategies against a baseline approach that uses the remaining data augmented with randomly selected samples instead of intelligently generated twins. The goal is to assess the robustness of the healing process and, specifically, the benefit of sophisticated twin generation when models are significantly impaired. The results indicate that, in most scenarios presented, employing twins related explicitly to these worst-affected samples provides better healing outcomes than randomly chosen samples.

The healing process proved effective regardless of whether the initial unlearning was performed via Fisher or Influence methods. While the starting accuracies differed significantly, the best healing outcomes reached similarly high levels, very close to the Gold Standard in both cases, even after a fast 1-epoch healing step. This demonstrates the robustness of fine-tuning based healing approaches in correcting degradations coming from different approximate unlearning techniques.

 \subsection{Key Takeaways}

The experiments do not cover all the cases  
but they confirm certain assumptions and pave the way for a more detailed analysis. 
The most important conclusions from the experiments:

\begin{itemize}[topsep=0pt]
\item The adversary models are significantly different. The experiments show that the parameter-aware adversary is much stronger than the adversary in the blind model. It should be noted that the adversary in the strong model can potentially use more other attacks, which can lead to even more effective results. 
\item The healing procedure can significantly protect against adversarial unlearn in selected settings, especially for exact unlearn methods. According to the experiments carried out, healing is acceptable in terms of resource consumption. 
\item Approximated unlearning is very sensitive to adversarial unlearning, even in the blind model. Even a few unlearned procedures can completely destroy the model. 
\item Exact unlearning in the blind adversary model is, of course, very resistant to adversarial unlearning. However, it is difficult to assess the resistance to stronger adversary classes.

\end{itemize}

\section{Conclusions and Future Work }\label{conclusions}

This paper formulated observations on the adversarial unlearning risks and presented a new, generic approach that can mitigate, to some extent, the negative effects in chosen scenarios.
Our contribution is not a complete work but a starting point for a deeper assessment of threats from unlearning for different adversary models.
In addition, a promising direction seems to be the extension of healing as a remedy for dangers related to various adversary settings.


Let us formulate the 
further work directions which are 
critical to protecting machine unlearning from 
malicious adversaries.

\begin{itemize}[topsep=0pt]

\item Analytical assessment of the risks of malicious unlearning, i.e., assessing what an adversary can do in a specific pattern classification model. Mainly what the limitations are in lowering the accuracy for each class of adversary defined in sec.~\ref{attack}. 
What is the impact of benign unlearn requests from regular users on adversarial attacks when unlearn requests from different users are blended?   

\item Finding better methods to assess the similarity of spare elements that would provide better performance.

\item Studying analogies for problems other than classification, particularly generative models. In contrast to classification, assessing the model's performance is non-obvious for many models. 

\item Finding a minimal set of spare elements that provide some resistance to malicious unlearning for various models of the adversary. 

\item It is well-known that many classes of deep models use training data differently. Some are memorized, while others are rather generalized (see, for example, \cite{DBLP:conf/stoc/Feldman20}). Intuitively, we do not need to keep too many generalized elements. This paves the way for creating a better optimized set of spare items.

\item Investigating the relationship between healing and regular unlearning. The first approach is not essentially unlearned in the sense of metrics on the forgetting quality (the idea is to use a scenario where unlearning is apparent, and it is to fulfill formal requirements and not a real removal of knowledge from the model).

\item Group healing: removing a certain subset of $D_r \subset D$ can be replaced by another set $D' \subset  D_S$. Intuitively, this approach may reduce the size of $D_S$ for the case if $D_r$ is significantly large. 
\end{itemize}








\newpage

\bibliography{main}

\begin{thebibliography}{41}
\providecommand{\natexlab}[1]{#1}
\providecommand{\url}[1]{\texttt{#1}}
\expandafter\ifx\csname urlstyle\endcsname\relax
  \providecommand{\doi}[1]{doi: #1}\else
  \providecommand{\doi}{doi: \begingroup \urlstyle{rm}\Url}\fi

\bibitem[Agarwal et~al.(2016)Agarwal, Bullins, and Hazan]{Agarwal2016SecondOrderSO}
N.~Agarwal, B.~Bullins, and E.~Hazan.
\newblock Second-order stochastic optimization for machine learning in linear time.
\newblock \emph{J. Mach. Learn. Res.}, 18:\penalty0 116:1--116:40, 2016.

\bibitem[Bourtoule et~al.(2021)Bourtoule, Chandrasekaran, Choquette-Choo, Jia, Travers, Zhang, Lie, and Papernot]{Bourtoule:2021}
L.~Bourtoule, V.~Chandrasekaran, C.~A. Choquette-Choo, H.~Jia, A.~Travers, B.~Zhang, D.~Lie, and N.~Papernot.
\newblock Machine unlearning.
\newblock In \emph{2021 IEEE Symposium on Security and Privacy (SP)}, pages 141--159, 2021.
\newblock \doi{10.1109/SP40001.2021.00019}.

\bibitem[{California State Legislature}(2018)]{CCPA}
{California State Legislature}.
\newblock California consumer privacy act of 2018, 2018.
\newblock \newline\url{https://leginfo.legislature.ca.gov/faces/billTextClient.xhtml?bill_id=201720180AB375 }.

\bibitem[Cao et~al.(2023)Cao, Jia, Zhang, and Gong]{DBLP:conf/sp/CaoJZG23}
X.~Cao, J.~Jia, Z.~Zhang, and N.~Z. Gong.
\newblock Fedrecover: Recovering from poisoning attacks in federated learning using historical information.
\newblock In \emph{44th {IEEE} Symposium on Security and Privacy, {SP} 2023, San Francisco, CA, USA, May 21-25, 2023}, pages 1366--1383. {IEEE}, 2023.

\bibitem[Chen et~al.(2025)Chen, Li, Zhao, and Huai]{DBLP:journals/aim/ChenLZH25}
A.~Chen, Y.~Li, C.~Zhao, and M.~Huai.
\newblock A survey of security and privacy issues of machine unlearning.
\newblock \emph{{AI} Mag.}, 46\penalty0 (1), 2025.

\bibitem[Chen et~al.(2024)Chen, Yang, Xiong, Bai, Hu, Hao, Feng, Zhou, Wu, and Liu]{Chen:2023}
R.~Chen, J.~Yang, H.~Xiong, J.~Bai, T.~Hu, J.~Hao, Y.~Feng, J.~T. Zhou, J.~Wu, and Z.~Liu.
\newblock Fast model debias with machine unlearning.
\newblock In \emph{Proceedings of the 37th International Conference on Neural Information Processing Systems}, NIPS '23, Red Hook, NY, USA, 2024. Curran Associates Inc.

\bibitem[Choi et~al.(2020)Choi, Uh, Yoo, and Ha]{choi2020starganv2}
Y.~Choi, Y.~Uh, J.~Yoo, and J.-W. Ha.
\newblock Stargan v2: Diverse image synthesis for multiple domains.
\newblock In \emph{Proceedings of the IEEE Conference on Computer Vision and Pattern Recognition}, 2020.

\bibitem[Chundawat et~al.(2023)Chundawat, Tarun, Mandal, and Kankanhalli]{DBLP:journals/tifs/ChundawatTMK23}
V.~S. Chundawat, A.~K. Tarun, M.~Mandal, and M.~S. Kankanhalli.
\newblock Zero-shot machine unlearning.
\newblock \emph{{IEEE} Trans. Inf. Forensics Secur.}, 18:\penalty0 2345--2354, 2023.
\newblock \doi{10.1109/TIFS.2023.3265506}.
\newblock URL \url{https://doi.org/10.1109/TIFS.2023.3265506}.

\bibitem[Cin{\`{a}} et~al.(2024)Cin{\`{a}}, Grosse, Demontis, Biggio, Roli, and Pelillo]{SecSurv}
A.~E. Cin{\`{a}}, K.~Grosse, A.~Demontis, B.~Biggio, F.~Roli, and M.~Pelillo.
\newblock Machine learning security against data poisoning: Are we there yet?
\newblock \emph{Computer}, 57\penalty0 (3):\penalty0 26--34, 2024.
\newblock \doi{10.1109/MC.2023.3299572}.
\newblock URL \url{https://doi.org/10.1109/MC.2023.3299572}.

\bibitem[{Council of European Union}(2014)]{eu-269-2014}
{Council of European Union}.
\newblock Council regulation ({EU}) no 269/2014, 2014.
\newblock \newline\url{http://eur-lex.europa.eu/legal-content/EN/TXT/?qid=1416170084502&uri=CELEX:32014R0269}.

\bibitem[Dhasade et~al.(2024)Dhasade, Ding, Guo, Kermarrec, de~Vos, and Wu]{DBLP:conf/middleware/DhasadeD0KVW24}
A.~Dhasade, Y.~Ding, S.~Guo, A.~Kermarrec, M.~de~Vos, and L.~Wu.
\newblock Quickdrop: Efficient federated unlearning via synthetic data generation.
\newblock In J.~Cao, Z.~Jin, V.~Schiavoni, and J.~Edinger, editors, \emph{Proceedings of the 25th International Middleware Conference, {MIDDLEWARE} 2024, Hong Kong, SAR, China, December 2-6, 2024}, pages 266--278. {ACM}, 2024.

\bibitem[Di et~al.(2023)Di, Douglas, Acharya, Kamath, and Sekhari]{DBLP:conf/nips/DiDA0S23}
J.~Z. Di, J.~Douglas, J.~Acharya, G.~Kamath, and A.~Sekhari.
\newblock Hidden poison: Machine unlearning enables camouflaged poisoning attacks.
\newblock In \emph{Advances in Neural Information Processing Systems 36: Annual Conference on Neural Information Processing Systems 2023, NeurIPS 2023, New Orleans, LA, USA, December 10 - 16, 2023}, 2023.

\bibitem[Di et~al.(2025)Di, Yu, Vorobeychik, and Liu]{di2025adversarial}
Z.~Di, S.~Yu, Y.~Vorobeychik, and Y.~Liu.
\newblock Adversarial machine unlearning.
\newblock In \emph{The Thirteenth International Conference on Learning Representations}, 2025.
\newblock URL \url{https://openreview.net/forum?id=swWF948IiC}.

\bibitem[Feldman(2020)]{DBLP:conf/stoc/Feldman20}
V.~Feldman.
\newblock Does learning require memorization? a short tale about a long tail.
\newblock In K.~Makarychev, Y.~Makarychev, M.~Tulsiani, G.~Kamath, and J.~Chuzhoy, editors, \emph{Proceedings of the 52nd Annual {ACM} {SIGACT} Symposium on Theory of Computing, {STOC} 2020, Chicago, IL, USA, June 22-26, 2020}, pages 954--959. {ACM}, 2020.

\bibitem[Golatkar et~al.(2020{\natexlab{a}})Golatkar, Achille, and Soatto]{DBLP:conf/cvpr/GolatkarAS20}
A.~Golatkar, A.~Achille, and S.~Soatto.
\newblock Eternal sunshine of the spotless net: Selective forgetting in deep networks.
\newblock In \emph{2020 {IEEE/CVF} Conference on Computer Vision and Pattern Recognition, {CVPR} 2020, Seattle, WA, USA, June 13-19, 2020}, pages 9301--9309. Computer Vision Foundation / {IEEE}, 2020{\natexlab{a}}.
\newblock \doi{10.1109/CVPR42600.2020.00932}.

\bibitem[Golatkar et~al.(2020{\natexlab{b}})Golatkar, Achille, and Soatto]{Golatkar:2020}
A.~Golatkar, A.~Achille, and S.~Soatto.
\newblock { Eternal Sunshine of the Spotless Net: Selective Forgetting in Deep Networks }.
\newblock In \emph{2020 IEEE/CVF Conference on Computer Vision and Pattern Recognition (CVPR)}, pages 9301--9309, Los Alamitos, CA, USA, June 2020{\natexlab{b}}. IEEE Computer Society.

\bibitem[Golatkar et~al.(2021)Golatkar, Achille, Ravichandran, Polito, and Soatto]{DBLP:conf/cvpr/GolatkarARPS21}
A.~Golatkar, A.~Achille, A.~Ravichandran, M.~Polito, and S.~Soatto.
\newblock Mixed-privacy forgetting in deep networks.
\newblock In \emph{{IEEE} Conference on Computer Vision and Pattern Recognition, {CVPR} 2021, virtual, June 19-25, 2021}, pages 792--801. Computer Vision Foundation / {IEEE}, 2021.
\newblock \doi{10.1109/CVPR46437.2021.00085}.

\bibitem[Graves et~al.(2020)Graves, Nagisetty, and Ganesh]{DBLP:journals/corr/abs-2010-10981}
L.~Graves, V.~Nagisetty, and V.~Ganesh.
\newblock Amnesiac machine learning.
\newblock \emph{CoRR}, abs/2010.10981, 2020.
\newblock URL \url{https://arxiv.org/abs/2010.10981}.

\bibitem[Guo et~al.(2023)Guo, Goldstein, Hannun, and van~der Maaten]{guo2023certifieddataremovalmachine}
C.~Guo, T.~Goldstein, A.~Hannun, and L.~van~der Maaten.
\newblock Certified data removal from machine learning models, 2023.
\newblock URL \url{https://arxiv.org/abs/1911.03030}.

\bibitem[Hu et~al.(2024)Hu, Wang, Chang, Zhong, Sun, Hao, Zhu, and Xue]{hu2024dutyforgetrightassured}
H.~Hu, S.~Wang, J.~Chang, H.~Zhong, R.~Sun, S.~Hao, H.~Zhu, and M.~Xue.
\newblock A duty to forget, a right to be assured? exposing vulnerabilities in machine unlearning services, 2024.
\newblock URL \url{https://arxiv.org/abs/2309.08230}.

\bibitem[Hu et~al.(2023)Hu, Wu, Li, Long, Garrido, Ge, Ding, Forsyth, Li, and Song]{DBLP:journals/corr/abs-2307-02106}
Y.~Hu, F.~Wu, Q.~Li, Y.~Long, G.~M. Garrido, C.~Ge, B.~Ding, D.~A. Forsyth, B.~Li, and D.~Song.
\newblock Sok: Privacy-preserving data synthesis.
\newblock \emph{CoRR}, abs/2307.02106, 2023.
\newblock \doi{10.48550/ARXIV.2307.02106}.
\newblock URL \url{https://doi.org/10.48550/arXiv.2307.02106}.

\bibitem[Huang et~al.(2024)Huang, Liu, Chua, Ghazi, Kamath, Kumar, Manurangsi, Nasr, Sinha, and Zhang]{huang2024unlearnburnadversarialmachine}
Y.~Huang, D.~Liu, L.~Chua, B.~Ghazi, P.~Kamath, R.~Kumar, P.~Manurangsi, M.~Nasr, A.~Sinha, and C.~Zhang.
\newblock Unlearn and burn: Adversarial machine unlearning requests destroy model accuracy, 2024.
\newblock URL \url{https://arxiv.org/abs/2410.09591}.

\bibitem[Jeong et~al.(2024)Jeong, Ma, and Houmansadr]{jeong2024sokchallengesopportunitiesfederated}
H.~Jeong, S.~Ma, and A.~Houmansadr.
\newblock Sok: Challenges and opportunities in federated unlearning, 2024.
\newblock URL \url{https://arxiv.org/abs/2403.02437}.

\bibitem[Koh and Liang(2020)]{koh2020understandingblackboxpredictionsinfluence}
P.~W. Koh and P.~Liang.
\newblock Understanding black-box predictions via influence functions, 2020.
\newblock URL \url{https://arxiv.org/abs/1703.04730}.

\bibitem[Krizhevsky and Hinton(2009)]{krizhevsky2009learning}
A.~Krizhevsky and G.~Hinton.
\newblock Learning multiple layers of features from tiny images.
\newblock Technical report, University of Toronto, 2009.

\bibitem[Lapuschkin et~al.(2019)Lapuschkin, W\"{a}ldchen, Binder, Montavon, Samek, and M\"{u}ller]{Lapuschkin2019}
S.~Lapuschkin, S.~W\"{a}ldchen, A.~Binder, G.~Montavon, W.~Samek, and K.-R. M\"{u}ller.
\newblock Unmasking {C}lever {H}ans predictors and assessing what machines really learn.
\newblock \emph{Nature Communications}, 10\penalty0 (1):\penalty0 1096, 2019.

\bibitem[Lecun et~al.(1998)Lecun, Bottou, Bengio, and Haffner]{Lecun1998}
Y.~Lecun, L.~Bottou, Y.~Bengio, and P.~Haffner.
\newblock Gradient-based learning applied to document recognition.
\newblock \emph{Proceedings of the IEEE}, 86\penalty0 (11):\penalty0 2278--2324, 1998.
\newblock \doi{10.1109/5.726791}.

\bibitem[Ma et~al.(2024)Ma, Zheng, Hu, Wang, Wang, Ba, Qin, and Ren]{ma2024releasingmalevolencebenevolencemenace}
B.~Ma, T.~Zheng, H.~Hu, D.~Wang, S.~Wang, Z.~Ba, Z.~Qin, and K.~Ren.
\newblock Releasing malevolence from benevolence: The menace of benign data on machine unlearning, 2024.
\newblock URL \url{https://arxiv.org/abs/2407.05112}.

\bibitem[Mahadevan and Mathioudakis(2021)]{mahadevan2021certifiablemachineunlearninglinear}
A.~Mahadevan and M.~Mathioudakis.
\newblock Certifiable machine unlearning for linear models, 2021.
\newblock URL \url{https://arxiv.org/abs/2106.15093}.

\bibitem[Marchant et~al.(2021)Marchant, Rubinstein, and Alfeld]{DBLP:journals/corr/abs-2109-08266}
N.~G. Marchant, B.~I.~P. Rubinstein, and S.~Alfeld.
\newblock Hard to forget: Poisoning attacks on certified machine unlearning.
\newblock \emph{CoRR}, abs/2109.08266, 2021.
\newblock URL \url{https://arxiv.org/abs/2109.08266}.

\bibitem[Martens(2010)]{10.5555/3104322.3104416}
J.~Martens.
\newblock Deep learning via hessian-free optimization.
\newblock In \emph{Proceedings of the 27th International Conference on International Conference on Machine Learning}, ICML'10, page 735–742, Madison, WI, USA, 2010. Omnipress.
\newblock ISBN 9781605589077.

\bibitem[Martens(2020)]{martens2020newinsightsperspectivesnatural}
J.~Martens.
\newblock New insights and perspectives on the natural gradient method, 2020.
\newblock URL \url{https://arxiv.org/abs/1412.1193}.

\bibitem[Martens and Grosse(2020)]{martens2020optimizingneuralnetworkskroneckerfactored}
J.~Martens and R.~Grosse.
\newblock Optimizing neural networks with kronecker-factored approximate curvature, 2020.
\newblock URL \url{https://arxiv.org/abs/1503.05671}.

\bibitem[Mercuri et~al.(2022)Mercuri, Khraishi, Okhrati, Batra, Hamill, Ghasempour, and Nowlan]{mercuri2022introductionmachineunlearning}
S.~Mercuri, R.~Khraishi, R.~Okhrati, D.~Batra, C.~Hamill, T.~Ghasempour, and A.~Nowlan.
\newblock An introduction to machine unlearning, 2022.
\newblock URL \url{https://arxiv.org/abs/2209.00939}.

\bibitem[Nguyen(2025)]{awesome}
T.~T. Nguyen.
\newblock {awesome-machine-unlearning}.
\newblock Online repository. GitHub, 2025.
\newblock Available: \url{https://github.com/tamlhp/awesome-machine-unlearning}.

\bibitem[Nguyen et~al.(2022)Nguyen, Huynh, Nguyen, Liew, Yin, and Nguyen]{nguyen2022surveymachineunlearning}
T.~T. Nguyen, T.~T. Huynh, P.~L. Nguyen, A.~W.-C. Liew, H.~Yin, and Q.~V.~H. Nguyen.
\newblock A survey of machine unlearning, 2022.
\newblock URL \url{https://arxiv.org/abs/2209.02299}.

\bibitem[Qian et~al.(2023)Qian, Zhao, Le, Ma, and Huai]{DBLP:conf/kdd/QianZLMH23}
W.~Qian, C.~Zhao, W.~Le, M.~Ma, and M.~Huai.
\newblock Towards understanding and enhancing robustness of deep learning models against malicious unlearning attacks.
\newblock In A.~K. Singh, Y.~Sun, L.~Akoglu, D.~Gunopulos, X.~Yan, R.~Kumar, F.~Ozcan, and J.~Ye, editors, \emph{Proceedings of the 29th {ACM} {SIGKDD} Conference on Knowledge Discovery and Data Mining, {KDD} 2023, Long Beach, CA, USA, August 6-10, 2023}, pages 1932--1942. {ACM}, 2023.

\bibitem[Shaik et~al.(2024)Shaik, Tao, Xie, Li, Zhu, and Li]{shaik2024exploringlandscapemachineunlearning}
T.~Shaik, X.~Tao, H.~Xie, L.~Li, X.~Zhu, and Q.~Li.
\newblock Exploring the landscape of machine unlearning: A comprehensive survey and taxonomy, 2024.
\newblock URL \url{https://arxiv.org/abs/2305.06360}.

\bibitem[Thudi et~al.(2022)Thudi, Jia, Shumailov, and Papernot]{DBLP:conf/uss/ThudiJSP22}
A.~Thudi, H.~Jia, I.~Shumailov, and N.~Papernot.
\newblock On the necessity of auditable algorithmic definitions for machine unlearning.
\newblock In K.~R.~B. Butler and K.~Thomas, editors, \emph{31st {USENIX} Security Symposium, {USENIX} Security 2022, Boston, MA, USA, August 10-12, 2022}, pages 4007--4022. {USENIX} Association, 2022.
\newblock URL \url{https://www.usenix.org/conference/usenixsecurity22/presentation/thudi}.

\bibitem[Wallis and Buvat(2022)]{CHmedicine}
D.~Wallis and I.~Buvat.
\newblock Clever hans effect found in a widely used brain tumour mri dataset.
\newblock \emph{Medical Image Analysis}, 77:\penalty0 102368, 2022.
\newblock ISSN 1361-8415.
\newblock \doi{https://doi.org/10.1016/j.media.2022.102368}.
\newblock URL \url{https://www.sciencedirect.com/science/article/pii/S1361841522000214}.

\bibitem[Xu et~al.(2023)Xu, Wu, Wang, and Jia]{Survey}
J.~Xu, Z.~Wu, C.~Wang, and X.~Jia.
\newblock Machine unlearning: Solutions and challenges.
\newblock \emph{CoRR}, abs/2308.07061, 2023.
\newblock \doi{10.48550/ARXIV.2308.07061}.
\newblock URL \url{https://doi.org/10.48550/arXiv.2308.07061}.

\end{thebibliography}


\end{document}